\documentclass[letterpaper, 10 pt, conference]{ieeeconf}  

\IEEEoverridecommandlockouts                              

\usepackage{hyperref}
\hypersetup{
    colorlinks=true,
    linkcolor=blue,
    filecolor=magenta,      
    urlcolor=blue,
    citecolor=blue,
}
\usepackage[dvipsnames]{xcolor}
\usepackage{booktabs}   
\usepackage{tabularx}  
\usepackage{amsmath}   
\usepackage{amssymb}
\usepackage{algorithm} 
\usepackage{algorithmic}
\usepackage{float}
\usepackage{multirow}
\usepackage{tikz}
\usetikzlibrary{shapes,arrows,calc,positioning}

\title{\LARGE \bf
ToPos: Automated Optimal Positioning on Topographic Manifolds using Constrained Geodesic Voronoi Decomposition
}

\author{ Rajesh Raveendran$^{1}$ Akseli Vanhamaa$^{2}$ Jaakko Suutala$^{1}$ Antti Tikanmäki$^{1}$  Juha Röning$^{1}$  
\thanks{$^{1}$Rajesh Raveendran, $^{1}$Jaakko Suutala, $^{1}$Antti Tikanmäki and $^{1}$Juha Röning are with Faculty of Information Technology and Electrical Engineering, Biomimetics and Intelligent Systems Group (BISG),
         University of Oulu, Pentti Kaiteran katu 1, 90570 Oulu, Finland
         {\tt\small rajesh.raveendran@oulu.fi}}%
 \thanks{$^{2}$Akseli Vanhamaa is Master’s student in Computer Science and Engineering, University of Oulu, Pentti Kaiteran katu 1, 90570 Oulu, Finland.
         {\tt\small}}%
 }
\begin{document}
\bstctlcite{IEEEexample:BSTcontrol}
\maketitle
\thispagestyle{empty}
\pagestyle{empty}
\begin{abstract}
Reliable autonomous mapping, environmental sampling, last-mile logistics, and infrastructure deployment depend on the optimal surface area-balanced distribution of Spatial Reference Sites (SRS). Conventional 2D Euclidean methods often fail in high-relief environments by neglecting topographic variations and physical obstructions. This leads to significant planimetric distortion, spatial clustering, and the placement of targets in inaccessible or shadowed regions, compromising both data integrity and operational safety. This paper introduces \textsc{ToPos}, an automated framework for TOPography-aware Optimal Sampling on topographic manifolds. We treat the terrain as a discrete 2-dimensional  manifold embedded in 3D Euclidean space and replace standard flat-map distances with non-Euclidean geodesic distances that follow the actual surface geometry. The point distribution is formulated as an optimization problem using a Constrained Geodesic Voronoi Decomposition, solved via a Riemannian Nesterov Accelerated Gradient (NAG) engine. Our approach restricts target locations to a feasible "safe zone," accounting for non-traversable slopes, vegetation, environmental occlusions, etc. Through evaluations on non-convex sinusoidal manifolds, we show that \textsc{ToPos} mitigates planimetric distortion by utilizing geodesic metrics. This approach results in a $\sim$74\% improvement in optimal surface area-balanced distribution, as measured by the \textit{coefficient of variation (CV)} of the Voronoi cell areas. The framework is architected as a  Geographic Information System (GIS)-ready micro-service to bolster the mentioned applications. 
\end{abstract}

\section{INTRODUCTION}
\subsection{Motivation}
In the context of autonomous drone surveying, the optimal surface area-balanced distribution of Ground Control Points (GCPs) is necessary to prevent geometric warping and minimize the Geometric Dilution of Precision (GDOP). Conventional 2D Euclidean tools induce spatial clustering in high-relief areas that compromises the structural correctness and data integrity of the final 3D model. \textsc{ToPos} provides an optimal solution for placing GCPs by enforcing surface area-balanced centroids across the topographic manifold. By distributing GCPs according to the intrinsic surface area rather than a distorted 2D projection, the framework maintains a spatial resolution and geometric weight that both directly and indirectly lowers GDOP and reduces the Root Mean Square Error (RMSE) of the mapping mission. 

Autonomous robots are considered for collecting samples from topographically complex environments periodically. In such scenarios, the selection of Ground Sampling Points (GSPs) must achieve an  optimally area-balanced distribution to maintain spatial homogeneity while strictly avoiding non-traversable areas or occluded regions where robots cannot travel. Methods that are based on standard 2D Euclidean partitioning often fail by placing these GSPs on vertical cliffs or within dense vegetation, leading to autonomous mission failure. \textsc{ToPos} addresses this by treating the environment as a discrete topographic manifold. By enforcing hard and soft manifold constraints, the \textsc{ToPos} optimization engine positions every GSP mathematically balanced according to the surface area while remaining within the mechanical, navigational, and energy envelope of the autonomous robots and maintaining spatial homogeneity. 

The limitations of 2D Euclidean partitioning are most acute in topographically extreme urban environments, such as San Francisco, where sidewalk gradients frequently exceed 20\% and vertical building envelopes create complex urban canyons. A robot attempting to reach a nearby staging point may encounter a non-traversable slope that was invisible in the planar coordinates, leading to mission failure or mechanical strain. This also causes more energy consumption than the initial planimetric estimate for the mission. Similarly, for autonomous drone delivery, standard flat-map siting neglects 3D building geometry and vertical obstacles that create GNSS shadows, compromising operational safety. \textsc{ToPos} addresses these challenges by treating the urban airspace and sidewalk paths as a discrete topographic manifold, guaranteeing that autonomous agents—whether on the ground or in the air—operate within their mechanical, energy, and signal-based safety envelopes.

\begin{figure}
\centering
\includegraphics[width=\columnwidth]{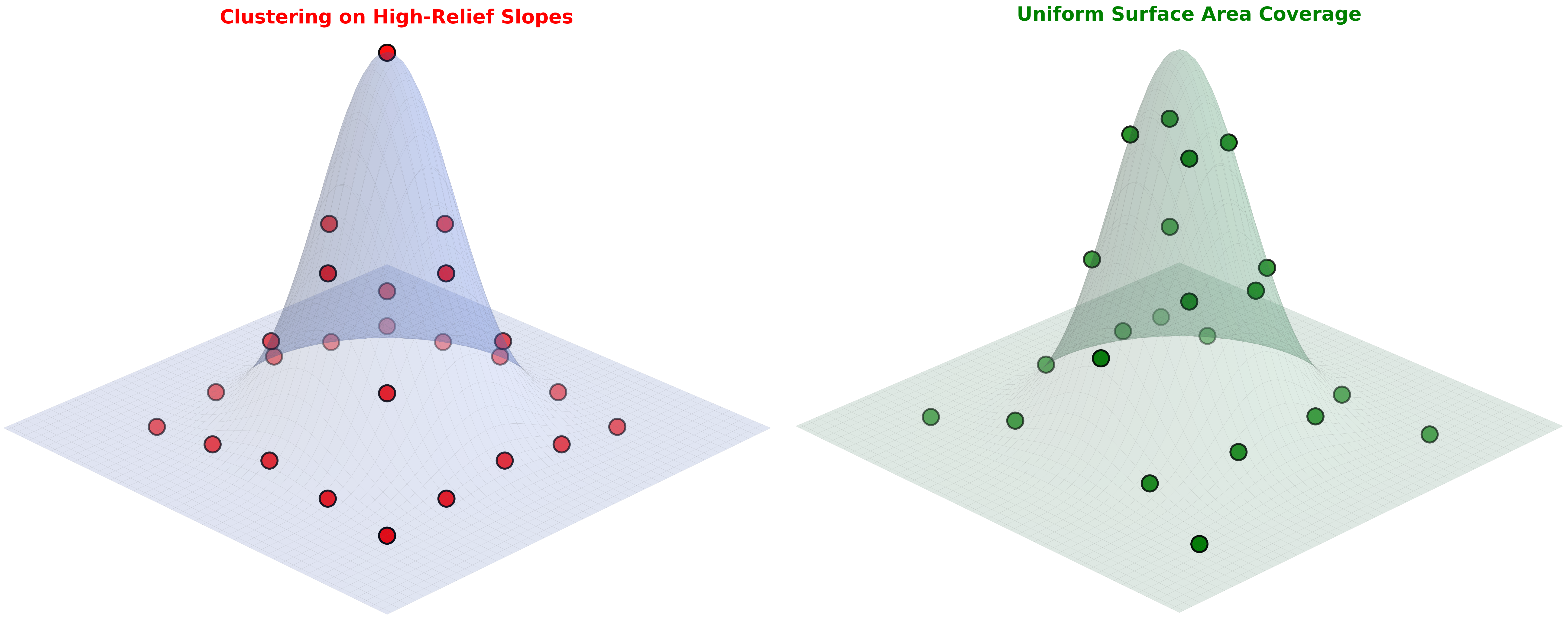}
\caption{Visualizing the \textbf{Metric Distortion Gap} and the \textsc{ToPos} solution. (Left) The planimetric 2D baseline fails to account for topographic relief, leading to \textbf{spatial clustering} and sparse coverage on steep slopes. (Right) The proposed \textsc{ToPos} framework achieves an \textbf{optimal area-balanced distribution} by optimizing directly on the manifold, ensuring uniform sampling density across high-relief terrain.}
\label{fig:intro_comparison}
\end{figure}

For telecommunication infrastructure siting, 2D projections fail to maintain metric uniformity along the manifold, misrepresenting the intrinsic surface area and geodesic distances each tower must cover. \textsc{ToPos} mitigates these risks by distributing infrastructure nodes according to the manifold's intrinsic geometry. Finally, in every positioning problem, whether used as GCPs, GSPs, a robotic staging hub, or a communication relay, it is mathematically optimized for maximum coverage and operational reliability in non-Euclidean environments. A visual comparison between the planimetric baseline and the \textsc{ToPos} area-balanced distribution is illustrated in Fig.~\ref{fig:intro_comparison}.
\subsection{Problem Statement}
The fundamental challenge in automated optimal positioning on topographic manifolds lies in the mathematical discrepancy between the 2D Euclidean projection used by standard GIS tools and the intrinsic 3D geometry of the physical environment. We define this as the \textit{Metric Distortion Gap}. Mathematically, this gap arises from the difference between the Euclidean metric $\delta_{ij}$ of a planar projection and the Riemannian metric $g_{ij}$ of the topographic surface. 

In high-relief regions, the local surface area element $d\mathcal{A} = \sqrt{\det(g)} \, du \, dv$ increases significantly relative to its planimetric footprint $du \, dv$. By neglecting this non-uniform metric density, conventional automated systems induce \textbf{spatial clustering} on the manifold $\mathcal{M}$, where the intrinsic geodesic distance $d_g(\mathbf{p}, \mathbf{q})$ between reference points is non-uniformly compressed. For AGVs and drones, this clustering directly compromises operational safety by placing targets in non-traversable voids, while simultaneously increasing the GDOP and degrading the structural integrity of the resulting 3D models. Therefore, there is a critical requirement for an automated engine that can optimize positioning directly on $\mathcal{M}$ using non-Euclidean geodesic metrics $d_g(\cdot, \cdot)$ and hard manifold constraints.

In summary, this paper has the following contributions:
\begin{itemize}
    \item[1)] \textit{A Robust Manifold-based Optimization Framework:} We introduce \textsc{ToPos}, an automated framework that utilizes discrete 2-manifold geometry to eliminate planimetric distortion in spatial reference sites distribution across high-relief terrains.
    \item[2)] \textit{Non-Euclidean Geodesic Partitioning:} We propose a Constrained Geodesic Voronoi Decomposition that employs intrinsic surface metrics rather than planar approximations, maintaining surface area-balanced distribution in topographically complex environments.
    \item[3)] \textit{Accelerated Riemannian Solver:} We implement a Riemannian Nesterov Accelerated Gradient (NAG) solver, achieving $\sim$2.4$\times$ faster convergence and a 74.17\% reduction in the \textit{Metric Distortion Gap} compared to standard Euclidean baseline iterations on non-convex topographic surfaces.


\end{itemize}
\section{Related Work} 
The challenge of area-balanced spatial distribution in complex environments spans across several domains, from classical computational geometry to modern autonomous robotics.
The mathematical foundation for optimal spatial partitioning is rooted in Centroidal Voronoi Tessellations (CVT). Traditionally, Lloyd's algorithm has been the standard for computing CVTs in Euclidean space \cite{lloyd_least_1982, du_centroidal_1999}. While highly effective for planar 2D domains, standard CVT solvers are inherently metric-blind to topographic relief. Previously there were attempts to adapt Euclidean-based CVT for manifolds \cite{liu_manifold_2016, wang_intrinsic_2015, khan_surface_2022}. As a fundamental challenge in distributed robotics, coverage control necessitates robust strategies for partitioning environments among multiple autonomous agents. By augmenting the classical centroidal update with a spatial density function, weighted CVT incorporates environmental demand directly into the tessellation’s geometric centers. Recently in \cite{gao_hybrid_2025} the author used a weighted CVT approach to identify optimal areas with high signal intensity in the context of heterogeneous sensor networks. Recent advancements in UAV tasking, such as the \textsc{SCOUT} framework \cite{peng_scout_2025}, have addressed spatiotemporal heterogeneity in demand during emergency response scenarios such as the 2025 Los Angeles wildfires. While \textsc{SCOUT} utilizes iterative gradient-based updates to enhance resource accessibility, it primarily operates on planimetric assumptions.

Precise GCP placement serves as the essential baseline for accurate 3D modeling of the environments \cite{triggs_bundle_2000}. In \cite{wangfei_selection_2009}, the authors used a weighted  Euclidean Voronoi to obtain GCPs in Image correction. Similarly, in another study \cite{guang_research_2011}, the author has addressed the impact of GCP distribution on geometric rectification using Voronoi. In both these approaches,  planimetric distortion was not considered, and the Euclidean-based Voronoi showed a lower RMSE of the model. In high-relief topography, the physical surveying of GCPs is often hindered by accessibility challenges, making the design of an effective placement strategy a prerequisite for 3D reconstruction in order to lower RMSE  \cite{cledat_mapping_2020}. When the environment's topology prevents the use of theoretically optimal locations for GCP placement, the framework must enable the user to re-evaluate operational parameters rapidly. A topographic manifold-aware framework will address this by providing a manifold-aware optimization engine that promises geometric precision and feasible solutions.

In one of the recent works \cite{meng_efficient_2023}, the author formulated a high-performance graph method based on Steiner point intersection to obtain geodesic distance to construct Voronoi on a manifold. However, these approaches don't consider constraints in robotics, and this fails to be a feasible solution. The Heat Method is one of the robust solvers for the Eikonal equation on triangular meshes \cite{crane_heat_2017} which helps to compute geodesics on a manifold. Other well-known methods, such as Dijkstra and Fast Marching, require calculations of distances from scratch for every iteration. In our work, we have used Heat Method, employed intrinsic geodesic geometry, and included the necessary constraints in robotics and drone applications. And it is also feasible to use it in any automated robotics framework. 

\section{Methodology}
\subsection{Manifold Representation}
We represent the topographic environment as a discrete 2-dimensional manifold $\mathcal{M}$ embedded in $\mathbb{R}^3$ Euclidean space. Formally, the surface is modeled as a simplicial complex (triangular mesh) $\mathcal{T} = \{V, F\}$, where $V \subset \mathbb{R}^3$ is the set of vertices and $F$ is the set of triangular faces, in which each vertex 
 $v_i$ is a point in 3D Euclidean space, represented by coordinates $(x_i, y_i, z_i) \in \mathbb{R}^3$. This embedding allows \textsc{ToPos} to compute intrinsic geodesic distances $d_g$ along the surface, thereby accounting for the true topographic relief that is otherwise lost in 2D Euclidean projections.
The feasible workspace $\mathcal{F} \subseteq \mathcal{M}$ is defined by hard manifold constraints. The objective of the \textsc{ToPos} framework is to minimize the \textit{Geodesic Voronoi Energy} $E(\mathcal{Z})$ shown in Eq. \eqref{eq:energy}, which measures the optimal surface area-balanced distribution of the seeds across the non-Euclidean manifold $\mathcal{M}$.

\begin{table*}[t] 
\centering
\caption{Comparison Between Standard 2D Euclidean Partitioning and the Proposed \textsc{ToPos} Framework}
\label{table:comparison}
\begin{tabularx}{\textwidth}{l X X} 
\toprule
\textbf{Feature} & 2D Euclidean Partitioning & \textbf{\textsc{ToPos} (Ours)} \\ 
\midrule
Manifold Space & Planar Euclidean ($\mathbb{R}^2$) & Non-Euclidean 2-Manifold ($\mathcal{M}$) \\ \addlinespace
Distance Metric & $L_2$ Norm (Straight-line distance) & Intrinsic Geodesic Metric ($d_g$) \\ \addlinespace
Area Measure & Planar Footprint Area & Topographic Surface Area element ($d\mathcal{A}$) \\ \addlinespace
Optimization Logic & Standard Lloyd's Iteration & Riemannian Nesterov Accelerated Gradient (NAG) \\ \addlinespace
Spatial Distribution & Spatial Clustering in High-Relief Zones &  Surface Area-Balanced Distribution Across the Manifold \\ \addlinespace
\bottomrule
\end{tabularx}
\end{table*}

\subsection{Geodesic Computation via the Heat Method}
\begin{figure}[H]
\centering
\includegraphics[width=\columnwidth]{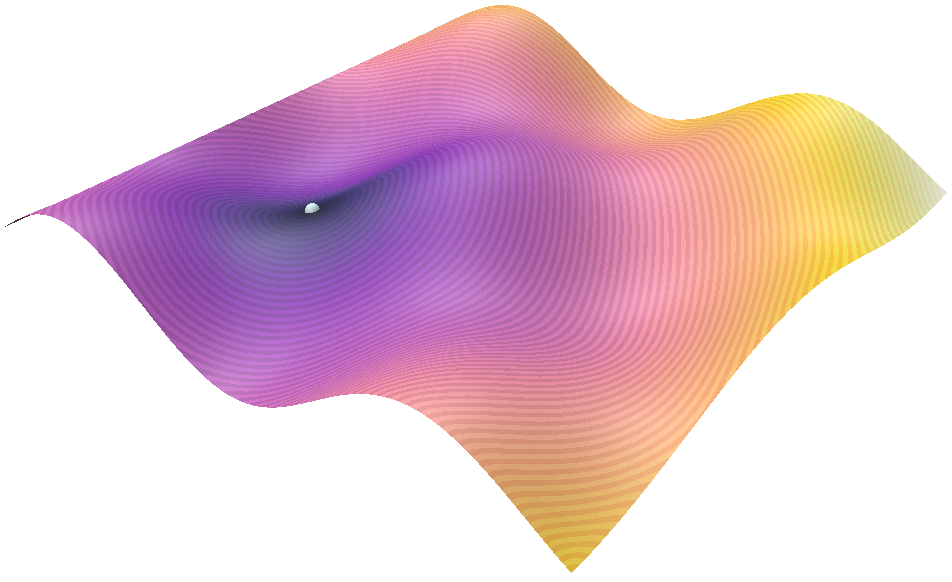}
\caption{Visualization of the \textbf{Heat Method} on the benchmark topographic manifold $\mathcal{M}$. The white \textbf{Eikonal isolines} represent equidistant geodesic contours emanating from a central \textbf{Spatial Reference Site} (white sphere). Note how the isolines deform to follow the intrinsic 3D curvature, providing the smooth, differentiable gradients required for \textsc{ToPos} optimization engine and voronoi decomposition.}
\label{fig:heat_method}
\end{figure}

To compute the intrinsic metric $d_g$ efficiently on high-resolution topographic meshes, \textsc{ToPos} employs the \textbf{Heat Method}. This approach approximates geodesic distances by solving a short-time heat diffusion problem on the simplicial complex $\mathcal{T}$. While the Fast Marching Method and Dijkstra's algorithm require a full re-propagation pass for every generator site update, the Heat Method treats each SRS as a heat source on the simplicial complex. By leveraging a precomputed Laplacian factorization, the engine can resolve the geodesic distance fields for all updated SRS locations nearly instantaneously. This provides the smooth gradients as shown in Fig. \ref{fig:heat_method}.
The computation of this method is performed in three discrete steps:

\begin{enumerate}
    \item \textit{Heat Diffusion:} We solve the discrete heat equation for a small time $t$ to obtain the heat distribution $u$ from a source at seed $\mathbf{z}_i$:
    \begin{equation}
    (I - t\Delta)u = \delta_{\mathbf{z}_i}
    \label{eq:heat_diffusion}
    \end{equation}
    where $\Delta$ is the discrete Laplace-Beltrami operator and $\delta_{\mathbf{z}_i}$ is the Dirac delta function at the seed location.
    
    \item \textit{Vector Field Evaluation:} The normalized negative gradient of the heat field is computed to obtain a unit vector field $X$ that points in the direction of the shortest paths:
    \begin{equation}
    X = -\frac{\nabla u}{|\nabla u|}
    \label{eq:vector_field}
    \end{equation}
    
    \item \textit{Poisson Integration:} Finally, the geodesic distance field $\phi \approx d_g$ is recovered by solving the Poisson equation:
    \begin{equation}
    \Delta \phi = \nabla \cdot X
    \label{eq:poisson}
    \end{equation}
\end{enumerate}

This integration allows \textsc{ToPos} to handle complex, high-relief geometries with minimal computational overhead, providing the necessary distance metrics for the subsequent Voronoi decomposition and Riemannian optimization.

\subsection{Geodesic Voronoi Decomposition} 
On a topographic manifold $\mathcal{M}$, a Voronoi cell $V_i$ associated with a generator (seed) $\mathbf{z}_i$ is defined as the set of all points $\mathbf{x} \in \mathcal{M}$ that are closer to $\mathbf{z}_i$ than to any other seed $\mathbf{z}_j$ according to the intrinsic geodesic metric $d_g$:
\begin{equation}
E(\mathcal{Z}) = \sum_{i=1}^{n} \int_{V_i \cap \mathcal{F}} d_g(\mathbf{x}, \mathbf{z}_i)^2 \, d\mathcal{A}
\label{eq:energy}
\end{equation}

where:
\begin{itemize}
    \item $V_i$ is the Voronoi cell associated with seed $\mathbf{z}_i$, defined by the set of all points closer to $\mathbf{z}_i$ than to any other seed $\mathbf{z}_j$ in the intrinsic metric: $V_i = \{ \mathbf{x} \in \mathcal{M} \mid d_g(\mathbf{x}, \mathbf{z}_i) \leq d_g(\mathbf{x}, \mathbf{z}_j), \, \forall j \neq i \}$.
    \item $d_g(\cdot, \cdot)$ is the intrinsic geodesic distance computed on the discrete manifold $\mathcal{M}$ via the \textit{Heat Method}.
    \item $d\mathcal{A}$ is the topographic surface area element, which mathematically accounts for the \textit{Metric Distortion Gap} by integrating the actual surface measure in high-relief regions.
    \item $\mathcal{F} \subseteq \mathcal{M}$ is the feasible manifold subset defined by hard operational constraints (e.g., local surface gradient $\nabla z < \theta_{\max}$ or visibility masks).
\end{itemize}

\subsection{Riemannian Optimization} 


To minimize $E(\mathcal{Z})$ on the manifold $\mathcal{M}$, we utilize the Exponential Map $\exp_{\mathbf{z}}(\mathbf{v})$ and the Logarithm Map $\log_{\mathbf{z}}(\mathbf{y})$. These operators facilitate the transition between the curved surface and the local Tangent Space $T_{\mathbf{z}}\mathcal{M}$. The \textsc{ToPos} Riemannian NAG update rules are formally expressed as:

\begin{equation}
\begin{aligned}
\mathbf{y}_k &= \exp_{\mathbf{z}_k}(\mu_k \mathbf{v}_k) \\
\mathbf{v}_{k+1} &= \Gamma_{\mathbf{y}_k \to \mathbf{z}_{k+1}} (\mu_k \mathbf{v}_k - \eta \text{grad}_{\mathcal{M}} E(\mathbf{y}_k)) \\
\mathbf{z}_{k+1} &= \exp_{\mathbf{y}_k}(-\eta \text{grad}_{\mathcal{M}} E(\mathbf{y}_k))
\end{aligned}
\label{eq:riemannian_nag}
\end{equation}

where $\text{grad}_{\mathcal{M}} E(\mathbf{z}) = -\log_{\mathbf{z}}(\mathbf{C}_i)$ represents the Riemannian gradient pointing towards the geodesic centroid $\mathbf{C}_i$, and $\Gamma$ denotes the \textit{parallel transport} of the velocity vector across the manifold. Fig.~\ref{Riemannian_Viz} illustrates the mapping between the topographic manifold $\mathcal{M}$ and the local tangent space $T_{\mathbf{z}}\mathcal{M}$ via the Logarithm and Exponential maps. This transition allows the \textsc{ToPos} engine to perform Euclidean-style Nesterov acceleration in the flat tangent plane before retracting the updated coordinates back to the intrinsic surface of the simplicial complex.

\usetikzlibrary{3d, perspective, arrows.meta}

\begin{figure}[t]
\centering
\begin{tikzpicture}[
    scale=2, 
    >=Stealth, 
    3d view={120}{25}, 
    grid line/.style={gray!30, very thin},
    manifold/.style={fill=blue!5, draw=blue!40, opacity=0.8, thick},
    tangent/.style={fill=gray!20, draw=black!60, opacity=0.7, thick}
]

    \draw[manifold] (-1.2,-1.2,0) 
        .. controls (0, -1.2, 0.5) and (1.2, -1.2, 0) .. (1.5, -1.2, -0.2)
        .. controls (1.5, 0, 0.3) and (1.5, 1.2, -0.2) .. (1.5, 1.5, 0)
        .. controls (0, 1.5, 0.6) and (-1.2, 1.5, 0) .. (-1.5, 1.5, -0.2)
        .. controls (-1.5, 0, 0.3) and (-1.5, -1.2, -0.2) .. cycle;

    \foreach \u in {-1,-0.5,0,0.5,1}{
        \draw[grid line] (\u,-1, 0.1) .. controls (\u,0,0.4) .. (\u,1.3,0.1);
        \draw[grid line] (-1.2,\u, 0.1) .. controls (0,\u,0.4) .. (1.3,\u,0.1);
    }
    \node[blue!70] at (1.3, -1, 0) {$\mathcal{M}$};

    \coordinate (Z) at (0,0,0.35); 
    \fill[black] (Z) circle (1pt) node[below left] {$\mathbf{z}_k$};

    \begin{scope}[shift={(0,0,0.35)}, plane origin={(0,0,0)}, plane x={(1,0,0)}, plane y={(0,1,0.3)}]
        \draw[tangent] (-0.8,-0.8) -- (0.8,-0.8) -- (0.8,0.8) -- (-0.8,0.8) -- cycle;
        
        \foreach \i in {-0.6,-0.3,0,0.3,0.6}{
            \draw[gray!40, ultra thin] (\i,-0.8) -- (\i,0.8);
            \draw[gray!40, ultra thin] (-0.8,\i) -- (0.8,\i);
        }
        
        \node[black!70] at (-0.5, -0.5, 0) {$T_{\mathbf{z}_k}\mathcal{M}$};
        
        \draw[->, red, very thick] (0,0) -- (0.6, 0.4) node[midway, above left] {$\mathbf{v}$};
    \end{scope}

    \coordinate (Y) at (1.1, 0.5, 0.1); 
    \fill[black] (Y) circle (1pt) node[right] {$\mathbf{y}$};

    \draw[blue, thick, dashed] (Z) .. controls (0.5, 0.4, 0.4) and (0.8, 0.4, 0.2) .. (Y);

    \draw[->, bend right=30, shorten >=2pt] (Y) to node[midway, right, font=\small] {$\log_{\mathbf{z}}$} (0.2, 0.15, 0.2);
    \draw[->, bend right=30, shorten <=2pt] (0.2, 0.21, 0.2) to node[midway, left, font=\small] {$\exp_{\mathbf{z}}$} (Y);

\end{tikzpicture}
\caption{Visualizing the \textbf{Riemannian Operators} on the topographic manifold. The Logarithm map $\log_{\mathbf{z}}$ pulls the 3D geodesic error into the local Euclidean tangent space $T_{\mathbf{z}_k}\mathcal{M}$, while the Exponential map $\exp_{\mathbf{z}}$ (realized as a retraction) maps the update vector $\mathbf{v}$ back onto the surface.}
\label{Riemannian_Viz}
\end{figure}
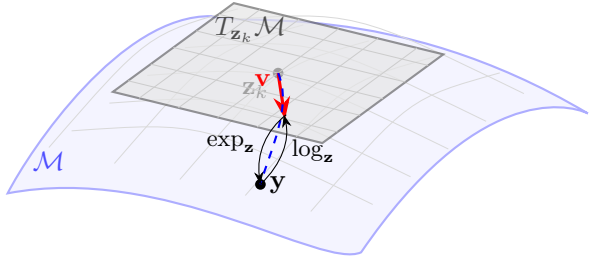

\subsection{Topographic Constraint Formulation}
To maintain operational safety, the optimization is governed by two manifold constraints: one that restricts where sites are placed and another that defines how the surface area $d\mathcal{A}$ is integrated.

\begin{enumerate}
    \item \textit{Hard Spatial Constraints (Feasible Set $\mathcal{F}$):} We define the feasible workspace $\mathcal{F} \subset \mathcal{M}$ as the subset of the manifold where physical placement of SRS is permitted. These constraints represent physical or administrative barriers where robotic access or equipment deployment is strictly prohibited. Formally, we require that for all generators $\mathbf{z}_i$, the condition $\mathbf{z}_i \in \mathcal{F}$ must be satisfied at every iteration. This is enforced through a manifold projection operator $\Pi_{\mathcal{F}}: \mathcal{M} \to \mathcal{F}$ that maps any infeasible update back to the nearest point on the boundary of the feasible set.

    \item \textit{Soft Metric Constraints (Metric Responsibility):} In contrast to total topological voids, certain restricted regions remain relevant to the mission's coverage objectives. These ``soft'' constraints retain their intrinsic surface area $d\mathcal{A}$ within the \textit{Geodesic Voronoi Energy} functional. While generators cannot occupy these zones, they must account for the area they encompass to minimize error in global surface area-balanced distribution. This forces the solver to optimize the distribution such that the surrounding feasible manifold perimeter $\mathcal{F}$ compensates for the service requirements of the non-traversable terrain.
\end{enumerate}

By differentiating between hard and soft, the \textsc{ToPos} methodology provides a flexible framework that maintains topographic accuracy while strictly adhering to the safety-critical boundaries of the real-world applications.

\subsection{The \textsc{ToPos} Optimization Algorithm}
The overall procedure for topography-aware optimal positioning is summarized in Algorithm \ref{alg:topos}. The engine iteratively alternates between geodesic computation, Voronoi partitioning, and Riemannian NAG updates until the energy functional $E(\mathcal{Z})$ reaches a stable equilibrium.

\begin{algorithm}[h]
\caption{The \textsc{ToPos} Framework}
\label{alg:topos}
\begin{algorithmic}[1]
\REQUIRE Mesh $\mathcal{T}$, Points $n$, Constraints $\mathcal{F}$, Iterations $K$
\STATE Initialize seeds $\mathcal{Z}_0 = \{\mathbf{z}_1, \dots, \mathbf{z}_n\} \subset \mathcal{F}$
\STATE Initialize velocity $\mathbf{v}_0 \leftarrow 0$
\FOR{$k = 1$ \TO $K$}
    \STATE \COMMENT{1. Geodesic Computation (Section III-B)}
    \STATE Compute $d_g(\mathbf{x}, \mathbf{z}_i)$ for all seeds via the \textbf{Heat Method}
    \STATE \COMMENT{2. Voronoi Partitioning (Section III-C)}
    \STATE Generate cells $V_i$ by assigning each $f \in F$ to the nearest seed
    \STATE \COMMENT{3. Riemannian NAG Update (Section III-D)}
    \STATE $\mathbf{y}_k \leftarrow \text{proj}_{\mathcal{F}}(\mathbf{z}_k + \mu_k \mathbf{v}_k)$ \hfill \textit{// Look-ahead}
    \STATE Compute $\nabla_{\mathcal{M}} E(\mathbf{y}_k)$ using surface area element $d\mathcal{A}$
    \STATE $\mathbf{v}_{k+1} \leftarrow \mu_k \mathbf{v}_k - \eta \nabla_{\mathcal{M}} E(\mathbf{y}_k)$ \hfill \textit{// Momentum}
    \STATE $\mathbf{z}_{k+1} \leftarrow \text{proj}_{\mathcal{F}}(\mathbf{z}_k + \mathbf{v}_{k+1})$ \hfill \textit{// Manifold Projection}
    \IF{$E(\mathcal{Z}_k)$ converges}
        \STATE \textbf{break}
    \ENDIF
\ENDFOR
\RETURN Optimized Seeds $\mathcal{Z}_{final}$
\end{algorithmic}
\end{algorithm}
\section{Implementation}

\subsection{Software Architecture and GIS Integration}
\begin{figure}[t]
\centering
\includegraphics[width=\columnwidth]{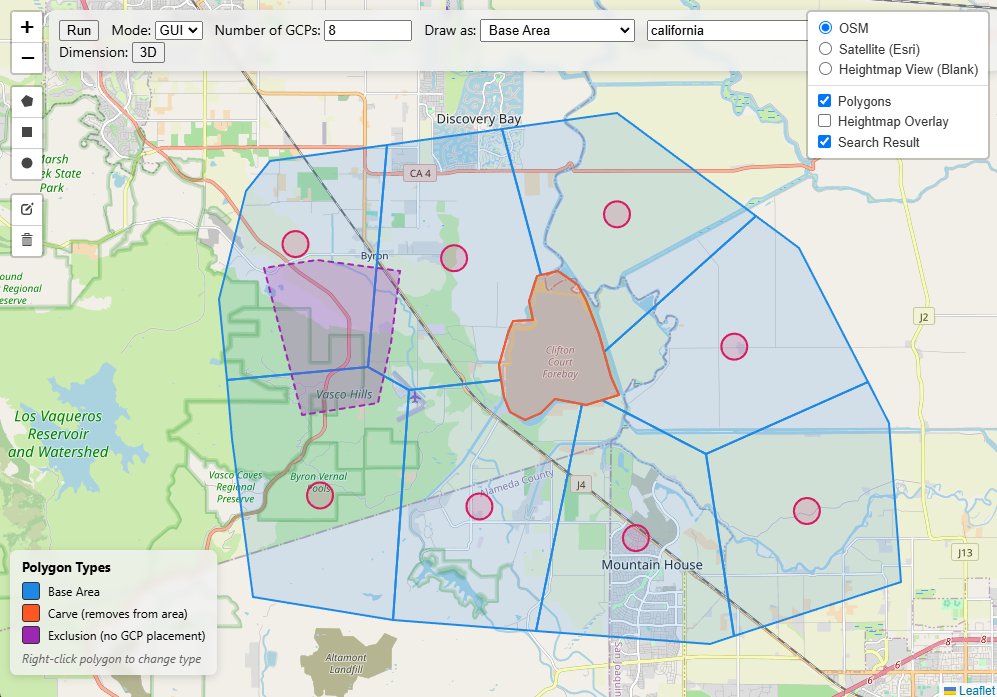}
\caption{The \textsc{ToPos} Graphical User Interface (GUI) demonstrating the integration of \textbf{Manifold-Aware} optimization with a standard GIS workflow. The interface highlights the \textbf{Base Area} (blue), \textbf{Operational Voids} (red regions, topologically carved out), and \textbf{Non-Traversable Zones} (purple regions, soft metric constraints). Note how the \textbf{Spatial Reference Sites} (red circles) are optimized to balance the surface area while strictly respecting the feasible boundary $\mathcal{F}$.}
\label{fig:interface_screenshot}
\end{figure}

The \textsc{ToPos} framework is designed as a modular micro-service architecture implemented in Python, leveraging \textit{PyQt6} for the desktop Graphical User Interface (GUI) and an embedded Leaflet-based map for area selection and real-time visualization as shown in Fig.~\ref{fig:interface_screenshot}. The implementation is divided into a GUI layer, an optimization core, and supporting utility modules. The system supports two optimization modes with a toggle switch in the GUI, allowing the operator to switch between a standard Euclidean (2D) and the proposed Manifold-Aware (3D) engine.
While the 2D mode utilizes \textit{Shapely} for planar geometry operations, the 3D mode transitions to an intrinsic Riemannian framework. We utilize the \textit{Geospatial Data Abstraction Library} (GDAL) to convert and reconstruct GeoTIFF heightmaps into simplicial complex $\mathcal{T}$ (triangular mesh). \textit{Potpourri3d} library is integrated to implement the Heat Method for intrinsic geodesic distance computation, while \textit{NumPy} is leveraged for the vectorized calculation of the 3D surface area $d\mathcal{A}$.

Input polygons are provided as WGS84 coordinates and transformed into a local Universal Transverse Mercator (UTM) projection using \textit{PyProj} library. This ensures that all geometric operations, including centroid calculation, and area computation, are performed in a consistent metric space. While the 2D mode treats GeoTIFF datasets solely as visual overlays and computation is performed in planar metric coordinates instead of directly in latitude and longitude, the 3D mode converts these rasters to generate a mesh, enabling the framework to transition from planar geometry to a rigorous Riemannian metric for high-relief terrain.

In addition, the software supports both GUI-based and file-based input to accommodate varying operational workflows. In GUI mode, the user selects the area of interest directly on the map interface and defines the number of SRSs to be generated. Alternatively, in file mode, these parameters are ingested via a YAML configuration file, facilitating automated batch processing. This modular structure ensures a strict separation between the user interaction layer and the core optimization logic, allowing \textsc{ToPos} to be deployed as either an interactive tool or a background GIS-ready micro-service.

\subsection{Constraint Implementation and Manifold Masking}
The framework utilizes a dual-masking architecture to enforce operational safety across both optimization modes. Operational Voids (Red regions) are treated as \textit{hard spatial constraints} and are topologically carved out from the manifold. By setting the local surface area $d\mathcal{A}$ to zero, these regions are effectively excluded from the optimization domain. In contrast, Non-Traversable Zones (Blue regions) act as \textit{soft metric constraints}; they retain their topographic area in the surface area-balanced distribution calculation but strictly prohibit physical generator placement.

This functional separation is critical for practical infrastructure deployment. For example, in the case of the spatial distribution of radio towers, a region containing a water body may be categorized as an Operational Void as it should not influence the final spatial distribution. Conversely, a roadway can be treated as a Non-Traversable Zone, where the area must remain part of the optimized signal coverage region even though a physical SRS cannot be placed within its boundaries.

The optimization engine processes the projected manifold geometry to compute the optimal placement of SRS. The resulting output consists of 1) a set of optimized 3D coordinates representing the site locations and 2) a topographic partition of the manifold into discrete Voronoi cells, where each cell represents a balanced portion of the total surface area. Upon completion of the optimization, these coordinates are transformed back into geographic coordinates (WGS84), archived in a timestamped CSV format, and rendered within the map interface for operational validation.\subsection{Riemannian Optimization Engine}
Under 3D mode, the Heat Method is central to this engine, which we utilize to solve the Eikonal equation across the simplicial complex $\mathcal{T}$. By precomputing the mesh Laplacian upon loading the data, the backend provides near-instantaneous geodesic distance queries, replacing the metric-blind Euclidean distances used in the 2D baseline.

In our discrete implementation, the Logarithm Map $\log_{\mathbf{z}}(\mathbf{y})$ is realized by evaluating the gradient of this geodesic distance field $\nabla \phi$ at each generator site. This operator "pulls" the topographic error into the local Euclidean tangent space $T_{\mathbf{z}}\mathcal{M}$, enabling the \textsc{ToPos} engine to perform momentum-based updates without the metric distortion inherent in 2D projections. Subsequently, the Exponential Map $\exp_{\mathbf{z}}(\mathbf{v})$ is implemented as a geodesic retraction onto the manifold, mapping the updated tangent vectors back to the nearest feasible vertices on the mesh. This "Map-Update-Retract" cycle ensures that the Nesterov look-ahead step remains strictly constrained to the topographic surface while achieving an accelerated $O(1/k^2)$ convergence rate.

\subsection{Experimental Setup}
\begin{figure}[t]
\centering
\includegraphics[width=\columnwidth, trim=0 20 0 20, clip]{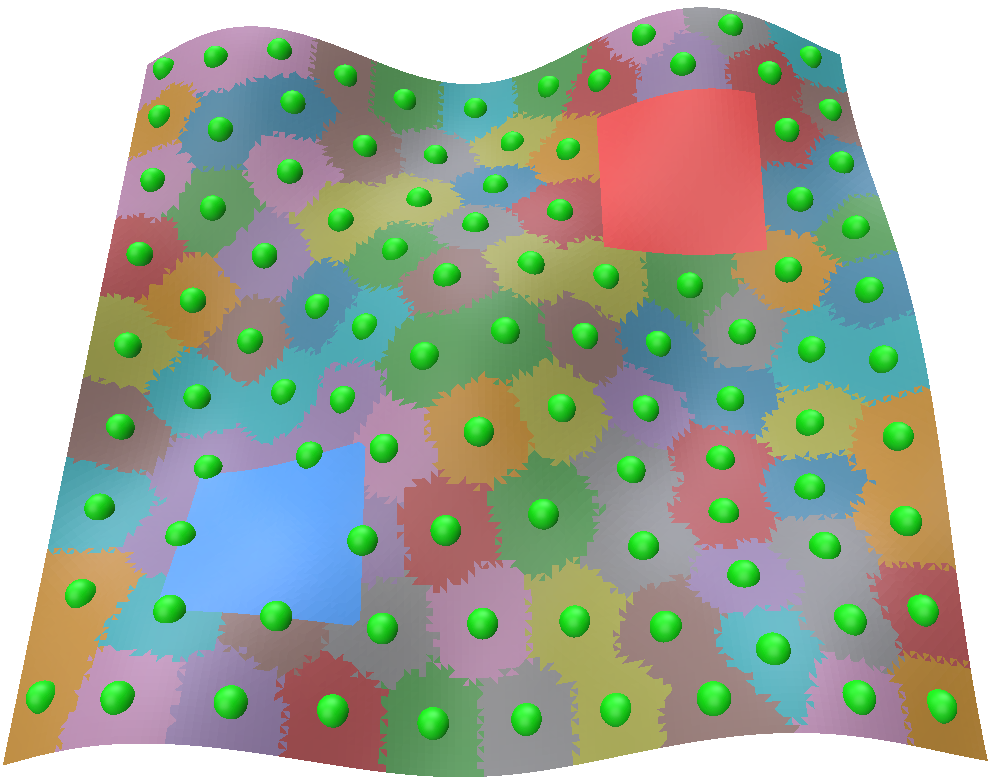}
\caption{Visualization of the \textsc{ToPos} experimental environment. The 16,900-vertex sinusoidal manifold is augmented with hard manifold constraints with generators(green spheres), including an \textbf{Operational Void} (red region) and a \textbf{Non-Traversable Zone} (blue region). Note how the \textsc{Geodesic Voronoi Cells} adapt to the topographic relief, ensuring that no generators ($n$ = 100) are placed within the forbidden zones.}
\label{fig:mesh_setup}
\end{figure}
To validate the proposed \textsc{ToPos} framework, we utilize a high-relief sinusoidal 2-manifold $\mathcal{M}$ defined over a $100 \times 100$ meter domain. This synthetic topography is characterized by a frequency-modulated elevation profile $z = \alpha \sin(x/\beta) \cos(y/\gamma)$, where the parameters are set to $\alpha = 7.0, \beta = 10.0$, and $\gamma = 15.0$. This induces significant planimetric distortion in standard Euclidean partitioning. This geometry serves as a rigorous non-convex baseline to evaluate the convergence and  surface area-balanced distribution of the \textsc{Riemannian NAG} solver compared to classical 2D Euclidean Lloyd's iterations.

As illustrated in Fig.~\ref{fig:mesh_setup}, the manifold is augmented with two distinct classes of constraints to evaluate the operational safety and coverage intelligence of the framework. We define an \textit{Operational Void} (red region) as a \textit{hard spatial constraint}, representing regions of total signal occlusion or environmental hazard. In this case, the manifold is topologically altered by setting the surface area $d\mathcal{A}$ to zero, effectively removing the region from the optimization functional. Conversely, the \textit{Non-Traversable Zone} (blue region) acts as a \textit{soft metric constraint}; while generator sites are strictly prohibited from physical placement within this zone to simulate robotic barriers, its topographic surface area $d\mathcal{A}$ remains active in the surface area-balanced calculation. Hence, \textsc{ToPos} maintains a uniform surface area-balance by optimizing the distribution to provide comprehensive coverage of the obstructed terrain, forcing generators to stabilize the reachable manifold $\mathcal{F}$ relative to the service needs of non-traversable regions.

The benchmark compares four strategies: (i) \textit{Standard 2D Lloyd's}, (ii) \textit{Adaptive Area-Balanced}, (iii) \textit{Riemannian Momentum}, and (iv) the proposed \textit{Riemannian NAG}. Each algorithm is tested across three point densities ($n \in \{10, 50, 100\}$) to assess scalability. To ensure statistical significance, we conduct 5 independent trials for each configuration, starting from randomized initial seed positions. Every trial is limited to a fixed computational budget of 100 iterations. We measure performance using the \textit{mean coefficient of variation} (CV) of Voronoi cell areas and the mean execution time per step to quantify the trade-off between geometric accuracy and computational overhead.
The performance of the proposed \textsc{ToPos} engine is evaluated based on its ability to minimize the variance of the intrinsic surface area across the topographic manifold. We quantify the \textit{Metric Distortion Gap} using the \textit{CV} of the geodesic Voronoi cell areas:

\begin{equation}
    CV = \frac{\sigma_{\mathcal{A}}}{\mu_{\mathcal{A}}} = \frac{\sqrt{\frac{1}{n} \sum_{i=1}^{n} (A_i - \bar{A})^2}}{\bar{A}}
\end{equation}

where:
\begin{itemize}
    \item $A_i$ represents the intrinsic surface area of the $i$-th geodesic Voronoi cell, calculated by the summation of the 3D triangular faces in the simplicial complex $\mathcal{T}$;
    \item $n$ is the total number of Voronoi cells;
    \item $\bar{A} = \frac{1}{n} \text{Area}(\mathcal{F})$ is the mean cell area calculated over the feasible manifold $\mathcal{F}$, where $\text{Area}(\mathcal{F}) = \sum A_i$.
\end{itemize}

Finally, the stability metric $\sigma$ is defined as the standard deviation across $N=5$ trials, computed as $\sigma = \sqrt{\frac{1}{N} \sum_{t=1}^{N} (CV_t - \mu)^2}$, where $CV_t$ is the final convergence value of trial $t$.
\subsection{Convergence Analysis}
\begin{figure}[H]
    \centering
    \includegraphics[width=\linewidth]{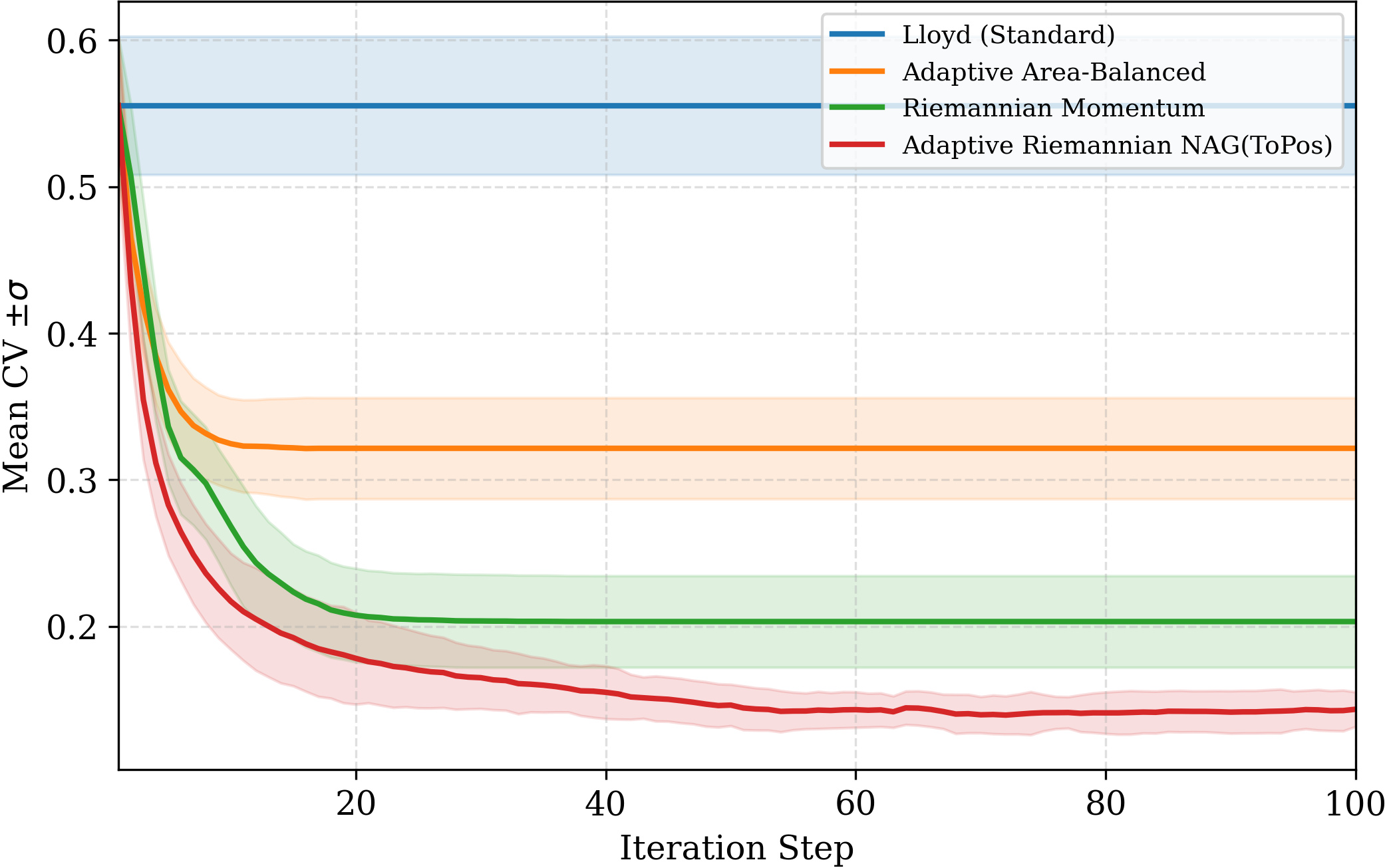}
    \vspace{-19pt}
    \caption{Convergence analysis (\textbf{$n$} = 100 Generators)}
    \label{fig:convergence}
\end{figure}

The convergence trajectories for a high-density configuration ($n=100$ generators) are illustrated in Fig.~\ref{fig:convergence} and summarized in Table~\ref{table:results_full}. While all manifold-aware methods outperform the Euclidean baselines, a distinct hierarchy in optimization efficiency is observed.

Initially, the standard 2D Lloyd baseline exhibits immediate asymptotic stagnation, remaining trapped in a high-energy state (CV $\approx 0.55$). This failure is a direct result of the \textit{Metric Distortion Gap} and the conflict between Euclidean updates and the manifold's feasible set $\mathcal{F}$. Because the solver is blind to the topographic relief, it repeatedly attempts to move toward centroids located within forbidden regions. Consequently, these updates are constrained to the feasible boundary, causing the baseline to stagnate as it lacks the intrinsic metric awareness required to account for the topographic surface area $d\mathcal{A}$.

The \textit{Adaptive Area-Balanced} method attempts to mitigate this stagnation by utilizing the Heat Method to resolve the intrinsic surface area $d\mathcal{A}$ and scaling Euclidean step sizes accordingly. However, as shown in the convergence logs, this heuristic approach lacks the geodesic awareness required to navigate the non-convex energy landscape effectively. While the \textit{Riemannian Momentum} method successfully incorporates the manifold's intrinsic metric, it lacks the proactive damping of the Nesterov look-ahead step, leading to suboptimal oscillations that prevent it from reaching the refined global equilibrium.

In contrast, the proposed \textsc{ToPos} engine (\textit{Adaptive Riemannian NAG}) achieves an accelerated descent, reaching a stable steady-state CV of $0.1434$. This represents a \textbf{74.17\%} improvement over the standard Euclidean-based Lloyd baseline and a \textbf{55.38\%} improvement over the Adaptive Area-Balanced method ($0.3214$) within the 100-iteration budget. Furthermore, the results demonstrate that the \textbf{Nesterov look-ahead step} prevents the generator sites from being trapped in the local minima induced by planimetric distortion, allowing \textsc{ToPos} to navigate the non-convex energy landscape toward a global equilibrium.

\begin{table}[t]
\centering
\caption{Quantitative Benchmarking Results across Generator Densities (100 Iterations)}
\label{table:results_full}
\setlength{\tabcolsep}{6pt}
\begin{tabularx}{\columnwidth}{@{}l l c c c@{}}
\toprule
\textbf{$n$} & \textbf{Algorithm} & \textbf{Mean CV} & \textbf{Std dev ($\sigma$)} & \textbf{Time/Step} \\ 
\midrule
\multirow{4}{*}{10} & Lloyd (Standard) & 0.4953 & 0.1643 & 79.8 ms \\
& Adapt. Area-Balanced & 0.2152 & 0.0597 & \textbf{78.9 ms} \\
& Riem. Momentum & 0.1426 & 0.0176 & 79.4 ms \\
& \textbf{ToPos (Proposed)} & \textbf{0.1541} & \textbf{0.0162} & 81.9 ms \\ 
\midrule
\multirow{4}{*}{50} & Lloyd (Standard) & 0.5396 & 0.0528 & 364.2 ms \\
& Adapt. Area-Balanced & 0.2750 & 0.0318 & \textbf{365.3 ms} \\
& Riem. Momentum & 0.1588 & 0.0181 & 366.9 ms \\
& \textbf{ToPos (Proposed)} & \textbf{0.1137} & \textbf{0.0172} & 382.1 ms \\ 
\midrule
\multirow{4}{*}{100} & Lloyd (Standard) & 0.5553 & 0.0423 & \textbf{727.0 ms} \\
& Adapt. Area-Balanced & 0.3214 & 0.0309 & 738.4 ms \\
& Riem. Momentum & 0.2032 & 0.0280 & 735.4 ms \\
& \textbf{ToPos (Proposed)} & \textbf{0.1434} & \textbf{0.0106} & 772.7 ms \\ 
\bottomrule
\end{tabularx}
\end{table}

The incorporation of the Exponential Map $\exp_{\mathbf{z}}(\mathbf{v})$ and Logarithm Map $\log_{\mathbf{z}}(\mathbf{y})$  on the Riemannian Manifold with the Nesterov look-ahead scheme allows the \textsc{ToPos} engine to achieve an accelerated $O(1/k^2)$ convergence rate. By performing momentum-based updates in the local tangent space and mapping them back to the manifold via the exponential map, the solver maintains the theoretical acceleration of Euclidean NAG while respecting the intrinsic geometry of the topographic manifold. In contrast, the Euclidean baseline exhibits the characteristic first-order $O(1/k)$ stagnation common in non-convex partitioning, as it lacks the manifold-aware mapping required to resolve the \textit{Metric Distortion Gap}.
 
The stability of the convergence is further evidenced by the shaded variance bands ($\pm \sigma$) in Fig.~\ref{fig:convergence}. Despite the non-convexity of the manifold, \textsc{ToPos} results in a \textbf{4.0$\times$ improvement in stochastic stability} ($\sigma = 0.0106$) compared to Euclidean methods. This result depicts that the look-ahead mechanism provides a more direct trajectory toward the geodesic centroid, ensuring that the final distribution of generators is highly robust to randomized initial seed placements.
\subsection{Geometric Stability}
\begin{figure}[H]
    \centering
    \includegraphics[width=\linewidth]{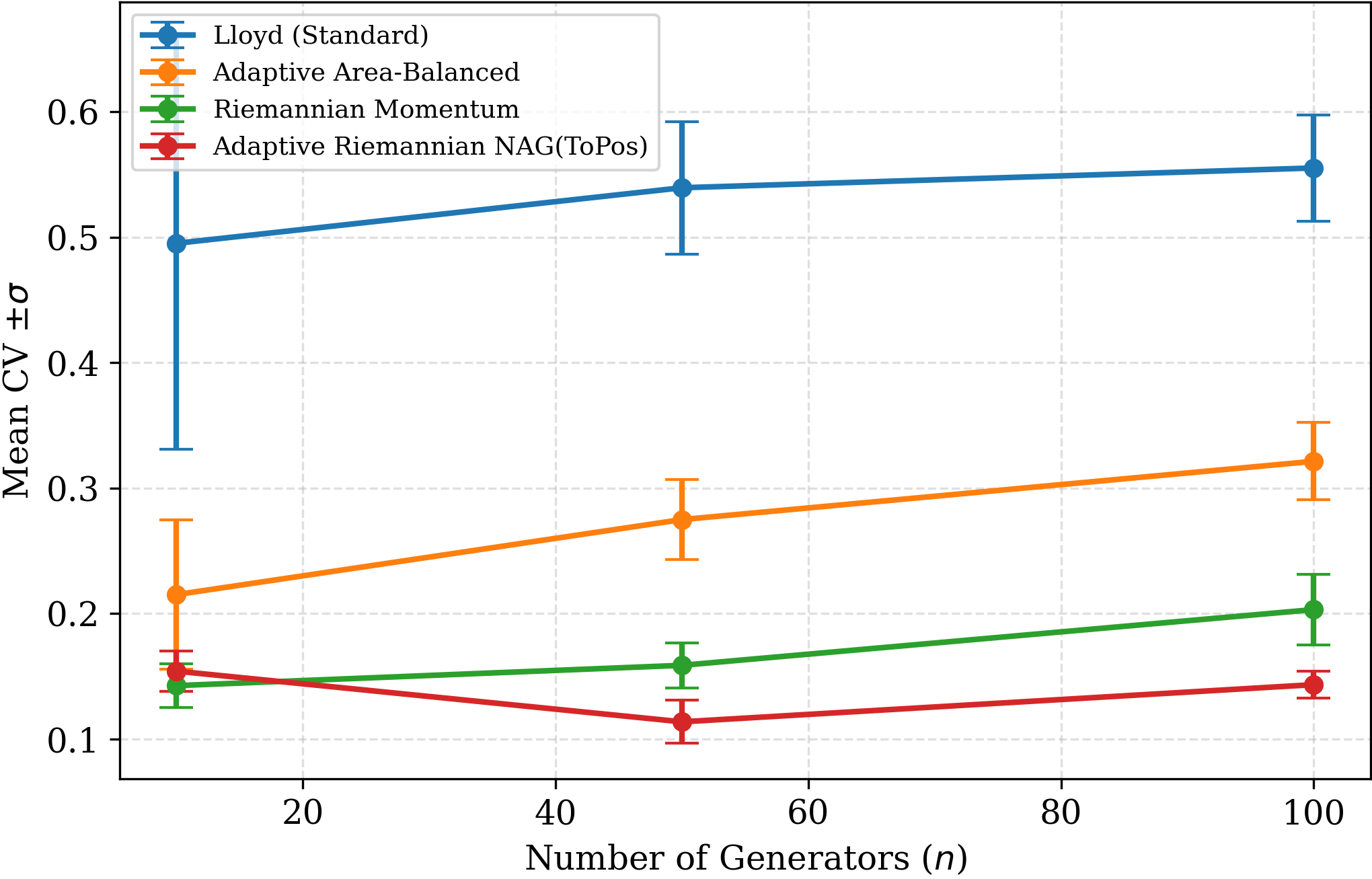}
    \vspace{-19pt}
    \caption{Geometric stability across generator densities}
    \label{fig:stability}
\end{figure}
The geometric stability of the proposed framework is evaluated by analyzing the sensitivity of the final spatial distribution to increasing generator densities ($n \in \{10, 50, 100\}$). As illustrated in Fig.~\ref{fig:stability}, \textsc{ToPos} demonstrates a superior invariance to scale, maintaining a consistent manifold-aware optimal surface area-balanced distribution even as the operational complexity and generator density grow. While standard Euclidean baselines exhibit severe spatial clustering, with the Lloyd's method failing at a Mean CV of $0.5553$ for $n=100$ the \textsc{ToPos} engine, it remains remarkably stable, settling at a refined CV of $0.1434$. Furthermore, such stochastic robustness is evidenced even when compared to the \textit{Adaptive Area-Balanced} method, which attempts to mitigate distortion through heuristic 2D scaling, with significantly lower standard deviation ($0.0106$ vs $0.0309$). These results show that while 2D adaptive heuristics may reduce mean error, they remain highly sensitive to initial conditions; in contrast, the \textsc{ToPos} engine demonstrates a repeatable and predictable area-balanced distribution that is essential for autonomous mission planning and infrastructure siting in complex 3D environments.

\subsection{Computational Efficiency}
\begin{figure}[t]
    \centering
    \includegraphics[width=\linewidth]{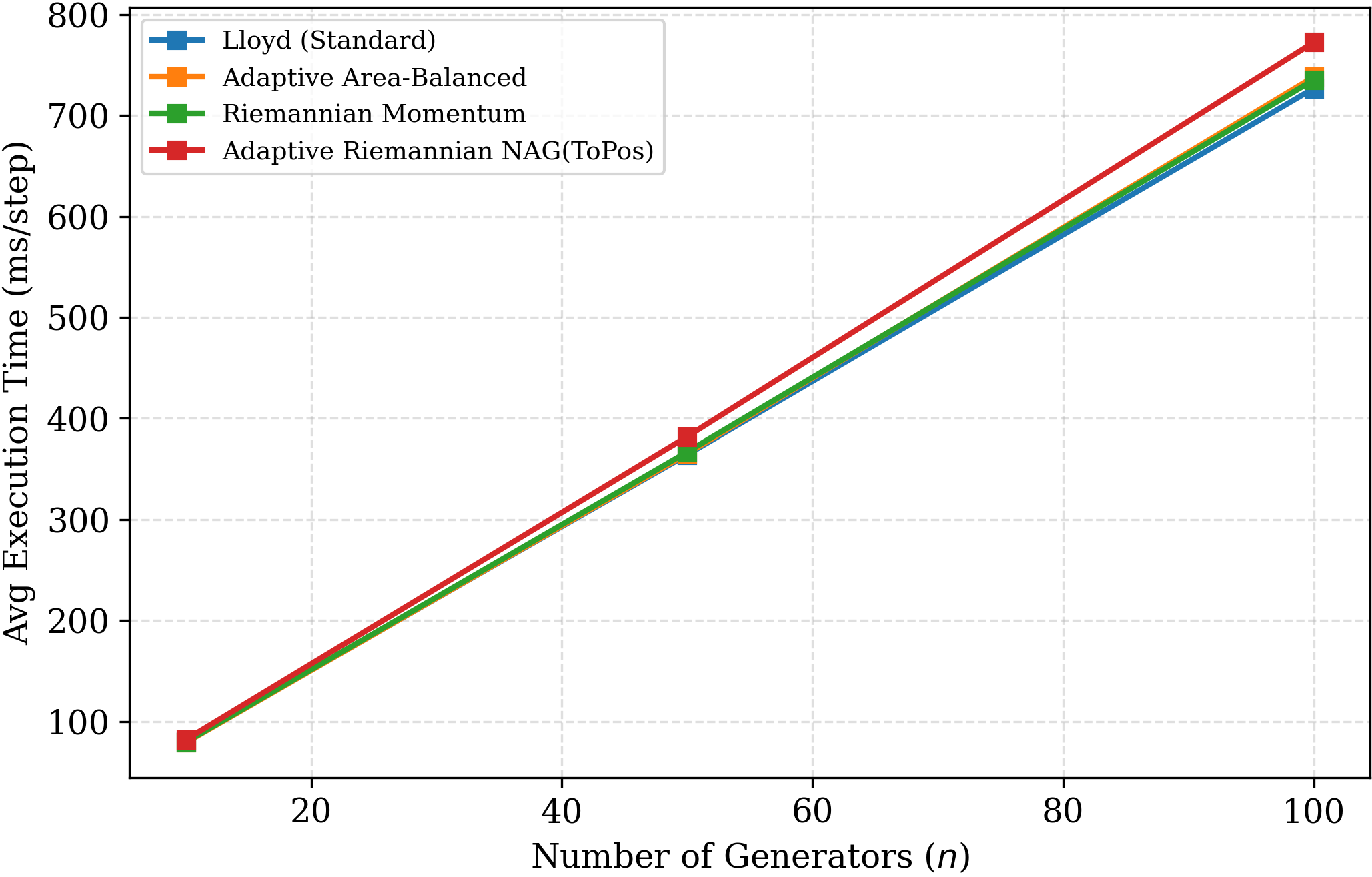}
    \vspace{-19pt}
    \caption{Computational efficiency analysis by comparing execution time across generator densities}
    \label{fig:computation}
\end{figure}
The computational cost per iteration is analyzed to evaluate the feasibility of the \textsc{ToPos} framework for real-time robotic deployment. As illustrated in Fig.~\ref{fig:computation}, the transition from a planimetric to a manifold-aware optimization domain introduces a negligible computational penalty. All tested algorithms exhibit a near-identical linear scaling $O(n)$ with respect to the number of generators. At the highest test density ($n=100$), the \textsc{ToPos} engine requires an average of $772.69$\,ms per iteration compared to $727.04$\,ms for the standard 2D Lloyd's baseline. This represents a marginal \textbf{6.28\% computational overhead}, which is a justified trade-off for the 74.17\% improvement in spatial surface area-balanced distribution. Furthermore, even when compared to the \textit{Adaptive Area-Balanced} method ($738.39$\,ms), the additional complexity of the Riemannian optimization results in only a \textbf{4.6\%} increase in execution time. Given that the execution time remains under $800$\,ms for a high-resolution {16,900-vertex} manifold, the system is well-suited for integration into autonomous mission planners or as a background GIS-ready micro-service. This shows that \textsc{ToPos} offers a superior accuracy-to-cost ratio, providing the necessary surface area-balanced distribution for safety-critical missions.

\section{Conclusion and Future Work}
We introduced ToPos, an automated framework that resolves the Metric Distortion Gap in spatial sampling by optimizing directly on discrete 2-manifolds. Integrating the Heat Method with a Riemannian Nesterov Accelerated Gradient engine allows ToPos to achieve a \textbf{74.17\%} reduction in CV cell area error and a $\approx$4.0$\times$ improvement in stochastic stability over standard planar Lloyd's baselines. This geometric precision is achieved with only a marginal \textbf{6.28\%} computational penalty, verifying its suitability for resource-constrained hardware. 

Future work will focus on extending the \textsc{ToPos} engine to support multiple specialized constraints. By broadening the constraint-handling architecture, we aim to make the framework viable for a wider range of industrial use cases, including large-scale infrastructure monitoring and heterogeneous multi-agent coordination in complex 3D environments. The core framework will be released as an open-source software.

\section*{Acknowledgment}
This research was fully supported by AUTONOMIC project funded by Interreg Aurora programme. The authors are highly grateful to Ville Pitkänen from the BISG, University of Oulu, for his profound insights during the mathematical discussions and valuable comments on the article.
\addtolength{\textheight}{-12cm}   






\bibliographystyle{IEEEtran}
\bibliography{references}

\end{document}